\documentclass[journal,twoside]{IEEEtran}
\usepackage{cite}
\usepackage[square, sort&compress, numbers]{natbib}
\usepackage{multirow}
\usepackage{graphicx}
\usepackage{amsmath,amssymb,amsfonts}
\allowdisplaybreaks[4]
\usepackage{algorithmic}
\usepackage{booktabs}
\usepackage{bm}
\usepackage{float}
\usepackage{bm}
\usepackage[linesnumbered,boxed,ruled,commentsnumbered]{algorithm2e}
\usepackage{textcomp}
\usepackage{threeparttable}
\def\BibTeX{{\rm B\kern-.05em{\sc i\kern-.025em b}\kern-.08em
    T\kern-.1667em\lower.7ex\hbox{E}\kern-.125emX}}
\begin{document}

\title{Confidence Estimation Transformer for Long-term Renewable Energy Forecasting in Reinforcement Learning-based Power Grid Dispatching}
\author{Xinhang Li, Zihao Li, Nan Yang, Zheng Yuan, Qinwen Wang, Yiying Yang, Yupeng Huang, Xuri Song\IEEEauthorrefmark{1}, Lei Li\IEEEauthorrefmark{1}, and Lin Zhang, \IEEEmembership{Member, IEEE}
\thanks{\IEEEauthorrefmark{1}Corresponding author}
\thanks{This work was supported by the National Natural Science Foundation of 
China (No.62176024) and Open Fund of Beijing Key Laboratory of Research and System Evaluation of Power Dispatching Automation Technology (China Electric Power Research Institute) (No.DZB51202101268).}
\thanks{X. Li, Z. Li, Z. Yuan, Q. Wang, Y. Yang, L. Li and L. Zhang are with Beijing University of Posts and Telecommunications, Beijing 100876, China (e-mail: \{lixinhang, lizihao, yuanzheng, wangqinwen, yyying, leili, zhanglin\}@bupt.edu.cn).}
\thanks{N. Yang, Y. Huang, X. Song is with Beijing Key Laboratory of Research and System Evaluation of Power Dispatching Automation Technology (China Electric Power Research Institute), Beijing 100192, China (e-mail: yn\_helen@163.com; 552324977@qq.com; songxuri@epri.sgcc.com.cn).}
}
\maketitle

\begin{abstract}
The expansion of renewable energy could help realizing the goals of peaking carbon dioxide emissions and carbon neutralization. Some existing grid dispatching methods integrating short-term renewable energy prediction and reinforcement learning (RL) have been proved to alleviate the adverse impact of energy fluctuations risk. However, these methods omit the long-term output prediction, which leads to stability and security problems on the optimal power flow. This paper proposes a confidence estimation Transformer for long-term renewable energy forecasting in reinforcement learning-based power grid dispatching (Conformer-RLpatching). Conformer-RLpatching predicts long-term active output of each renewable energy generator with an enhanced Transformer to boost the performance of hybrid energy grid dispatching. Furthermore, a confidence estimation method is proposed to reduce the prediction error of renewable energy. Meanwhile, a dispatching necessity evaluation mechanism is put forward to decide whether the active output of a generator needs to be adjusted. Experiments carried out on the SG-126 power grid simulator show that Conformer-RLpatching achieves great improvement over the second best algorithm DDPG in security score by 25.8$\%$ and achieves a better total reward compared with the golden medal team in the power grid dispatching competition sponsored by State Grid Corporation of China under the same simulation environment. Codes are outsourced in \emph{https://github.com/buptlxh/Conformer-RLpatching}.

\end{abstract}


\section{Introduction}
\label{sec:introduction}
\begin{figure*}[]
\centering
\centerline{\includegraphics[width=\textwidth,height=9.6cm]{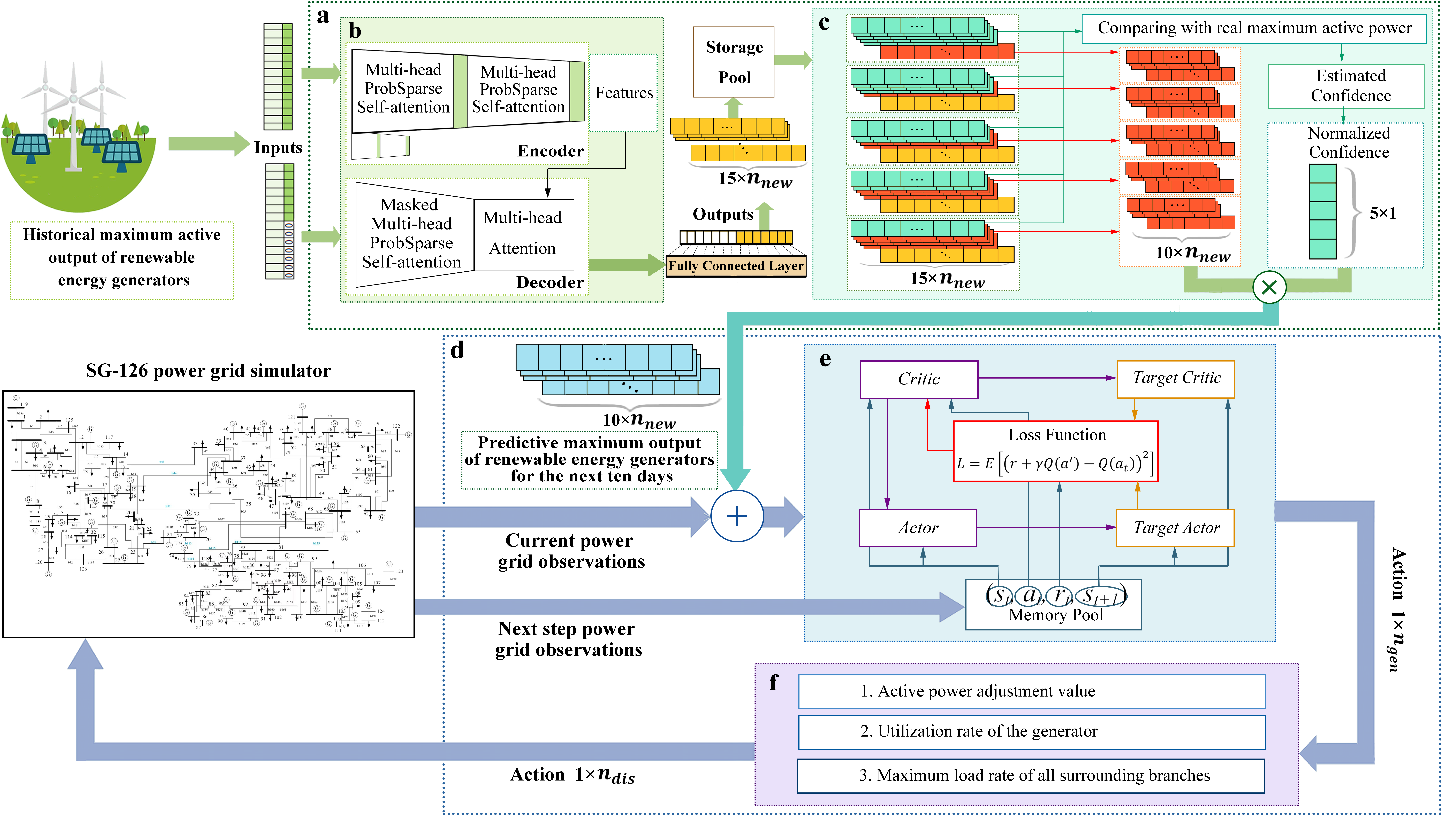}}
\caption{Architecture of Conformer-RLpatching. Conformer (\textbf{a}) obtains accurate renewable energy long-term prediction, and is divided into the enhanced Transformer-based renewable energy prediction model (\textbf{b}) and the confidence estimation method (\textbf{c}). RLpatching (\textbf{d}) provides an appropriate dispatching strategy, and contains the DDPG-based power flow optimization algorithm (\textbf{e}) and the dispatching necessity evaluation mechanism (\textbf{f}). 
}
\label{fig_1}
\end{figure*}
\IEEEPARstart{T}{he} randomness and volatility of renewable energy sources (RESs) have highlighted the pressing need to address stability and security concerns in power grid dispatching \cite{a1,a3}. Meanwhile, the utilization rate of renewable energy (URRE), as another performance index in addition to security and economy, brings new challenges to the security constrained economic dispatching (SCED) in the hybrid energy grid \cite{a5}. \cite{b1} presented a modified version of multi-objective differential evolution by incorporating wind power plant into the dynamic economic dispatch system. Aiming to minimize cost and restrict risks, a novel SCED model \cite{b2} and a chance-constrained economic dispatch model \cite{b3} were proposed for wind integrated hybrid power system. \cite{b4} implemented a short-term control algorithm to smoothen the power dispatching by improving min-max dispatching method. \cite{b5} proposed a look-ahead stochastic unit commitment model for robust optimal dispatching. However, all the above dispatching methods did not consider the volatility of RESs and URRE simultaneously.

The emerging artificial intelligence (AI) technology is increasingly applied to the hybrid energy grid dispatching \cite{a8}. In August 2021, State Grid Corporation of China (SGCC) and Baidu jointly sponsored State Grid Dispatching AI Innovation Competition, which is authoritative in smart grid.\footnote{\emph{https://aistudio.baidu.com/aistudio/competition/detail/111/0/introduction}} 
SGCC provided SG-126 power grid simulator for participating teams to realize multi-objective power grid dispatching and strive for the highest total reward. The total reward of the first ranked team is 510.09 among nearly 100 teams. In addition, advanced AI algorithms were put forward to further improve flexibility, controllability and observability of the hybrid grid \cite{a9}. An innovative dispatch optimization strategy with an uncertainty post-processing approach was proposed in \cite{c1}, and \cite{c2} adopted a learning-based technique to search the optimal joint control policy. \cite{c3} presented a manifold-learning-based Isomap algorithm to represent the low-dimensional hidden probabilistic structure of data. \cite{c4} proposed a learning-based decision-making framework for the economic energy dispatching based on historical sequences. However, the optimization objectives of \cite{c1,c2,c3,c4} did not involve URRE. \cite{c5} modeled the power dispatching as sequential decision-making and introduced Deep Reinforcement Learning (DRL). \cite{c6} applied deep deterministic policy gradient (DDPG) to microgrids with photovoltaic panels. \cite{c7} improved DRL to adaptively respond to the renewable energy fluctuations. The improved decision tree \cite{c8,c9} was also able to provide feasible and optimal dispatch decisions for microgrids. \cite{d1} developed a new renewable energy management system with short-term forecasting for hourly dispatching. \cite{d2} proposed a data-driven RL approach to relieve branch overload in large power systems.
However, all the above dispatching algorithms based on AI did not consider the long-term fluctuation of new energy and were only suitable for real-time or short-term dispatching.

In order to improve the accuracy of forecasting, some prediction algorithms have been proposed. \cite{d4,d5} adopted the method of historical data mining to analyze data characteristics for regional wind power prediction and wind power ramp prediction, respectively. \cite{d6} introduced small-world neural network to reduce wind power forecasting errors. \cite{d7,d8} both used hybrid models for high-precision day-ahead short-term photovoltaic output forecasting. Although Informer proposed in \cite{d9} has been verified to be able to accurately predict long-term electricity consuming load, it has not been applied to renewable energy prediction. 

In summary, the existing dispatching methods fail to achieve multi-objective dispatching for the hybrid power system under the long-term fluctuation of renewable energy. 
To address the problem, a confidence estimation Transformer for long-term renewable energy forecasting in reinforcement learning-based power grid dispatching (Conformer-RLpatching) is proposed shown in Fig. \ref{fig_1}.
In Conformer, the confidence estimation method weights the prediction results from the enhanced Transformer to reduce prediction error. RLpatching utilizes the long-term prediction from Conformer and the current power grid observations to provide an appropriate multi-objective dispatching strategy. The contributions of this paper are summarized as follows:

1. This paper proposes Conformer, an enhanced Transformer with a confidence estimation method, which provides accurate prediction to boost the performance of hybrid energy grid dispatching and abate the impact of renewable energy uncertainty.

2. This paper proposes Conformer-RLpatching to realize multi-objective dispatching under the long-term fluctuation of renewable energy. A dispatching necessity evaluation mechanism is put forward to reduce unnecessary dispatching, thereby improving the stability of the power grid.

3. This paper applies Conformer-RLpatching to the SG-126 power grid simulator provided by SGCC, bridging the ‘sim-to-real’ gap. Extensive experiments show that Conformer-RLpatching achieves great improvement over DDPG in security score by 25.8$\%$ and achieves remarkable total reward up to 527.32 superior to 510.09, the highest total reward in State Grid Dispatching AI Innovation Competition.\footnote{\emph{https://aistudio.baidu.com/aistudio/competition/detail/111/0/leaderboard}}


The rest of this paper is organized as follows. Section \ref{prediction} introduces the confidence estimation Transformer for long-term renewable energy forecasting. Section \ref{dispatching} presents the reinforcement learning-based dispatching framework and the dispatching process. Section \ref{results} gives the performance of proposed methods. Section \ref{conclu} draws the conclusion.

\section{Confidence Estimation Transformer for Long-term Renewable Energy Forecasting}\label{prediction}
This section mainly introduces two main components of Conformer, the enhanced Transformer-based renewable energy prediction model and the confidence estimation, which are responsible for the long-term prediction of renewable energy and the synthesis of prediction results, respectively. Section \ref{cp} introduces the specific process of Conformer.

\begin{figure*}[h]
\centering
\centerline{\includegraphics[width=0.98\textwidth,height=9.5cm]{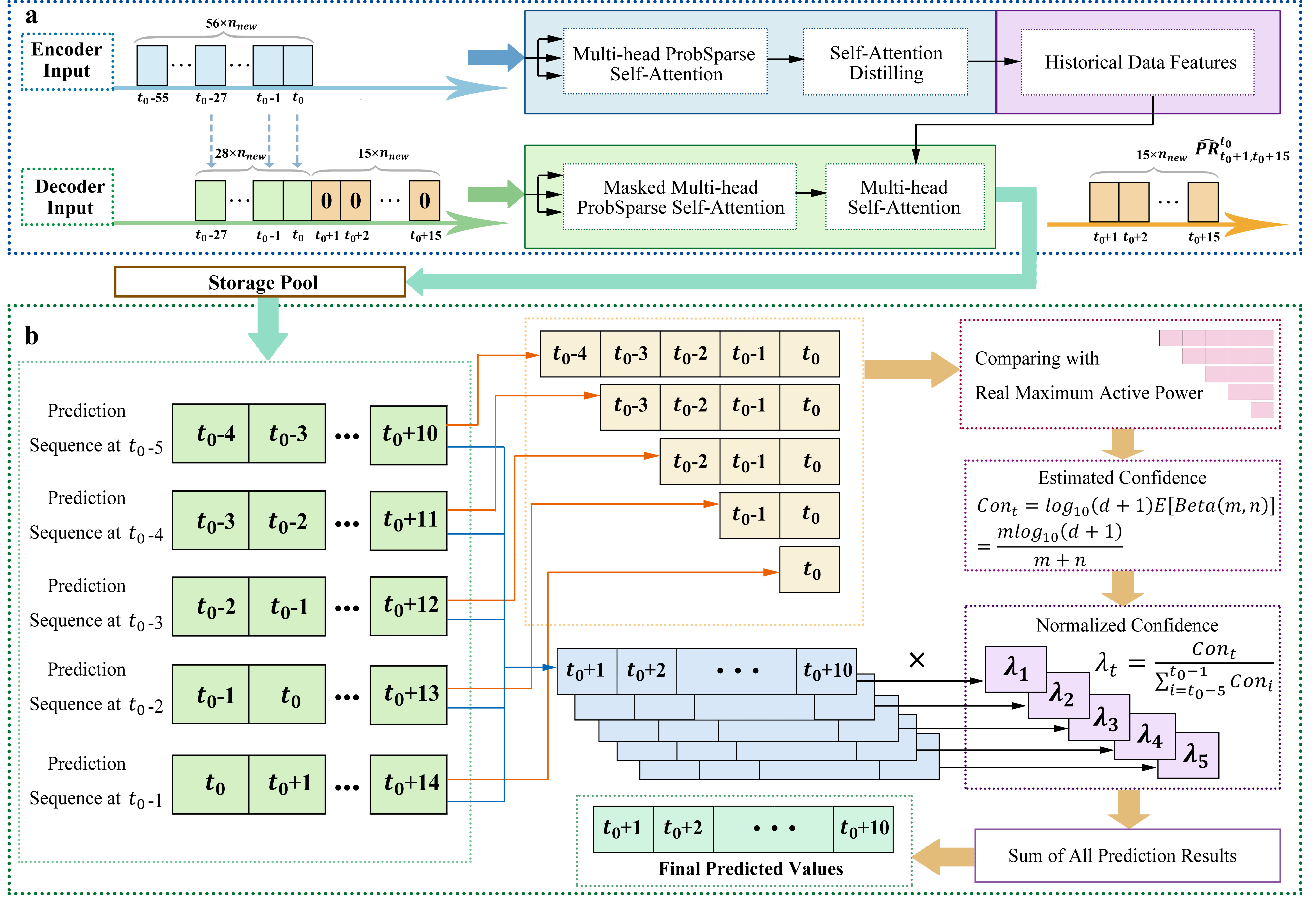}}
\caption{Flow Chart of Conformer. \textbf{a}, The enhanced Transformer-based renewable energy prediction model. \textbf{b}, The confidence estimation method.}
\label{fig_4}
\end{figure*}

\subsection{Enhanced Transformer for Renewable Energy Prediction}
In order to deal with renewable energy fluctuations caused by external factors in advance, this paper adopts the enhanced Transformer \cite{d9} to efficiently predict the long-term maximum active output of every renewable energy generator. The enhanced Transformer-based renewable energy prediction model, shown in Fig. \ref{fig_4} a, is divided into encoder and decoder. 


 For all renewable energy generators, ${\mathop { \textit{PR}}\nolimits_{{\mathop t\nolimits_1},{\mathop t\nolimits_2}}}$ represents real maximum active power from ${\mathop t\nolimits_1}$ to ${\mathop t\nolimits_2}$, and the sequence of the maximum active power predicted by the enhanced Transformer at time $t$ is noted as ${\mathop {\widehat {\textit{PR}}}\nolimits_{{\mathop t\nolimits_1},{\mathop t\nolimits_2}}^{t}}$. ${\mathop {\widehat {\textit{PR}}}\nolimits_{{t} + 1,{t} + 10}}$ represents the final prediction result for the next 10 days output by Conformer.

The encoder mainly carries out coding for known sequences and extracting features. Suppose that the current time is ${\mathop t\nolimits_0}$. In the enhanced Transformer, the input of encoder includes the real data in the past 56 days ${\mathop { \textit{PR}}\nolimits_{{\mathop t\nolimits_0} -55,{\mathop t\nolimits_0}}}$ 
and the corresponding four-dimension time code, whose sizes are $56 \times \mathop n\nolimits_{new}$ and $56\times4$, respectively. ProbSparse self-attention is adopted in the encoder, achieving $O(LlogL)$ time complexity and $O(LlogL)$ memory usage on dependency alignments. For tuple inputs, i.e, query $Q$, key $K$ and value $V$, ProbSparse self-attention allows each key to only attend to the $u$ dominant queries:
\begin{equation}
   {{\rm A}(Q, K, V) = Softmax(\frac{{\overline Q \mathop K\nolimits^T }}{{\sqrt d }})V}
\end{equation}
where $\overline Q$ is a sparse matrix of the same size as $Q$, and it only contains the Top-$u$ queries under the sparsity measurement. Meanwhile, self-attention distilling operation is introduced. Conv1D layers and Maxpooling layers are added after each ProbSparse self-attention layer to privilege dominating attention scores and help receiving long sequence input. 


The decoder is responsible for predicting. The input of the decoder includes historical data and corresponding time coding besides the sequence characteristics output by the encoder. The data section includes real data in the past 28 days ${\mathop { \textit{PR}}\nolimits_{{\mathop t\nolimits_0} -27,{\mathop t\nolimits_0}}}$ and that in the next 15 days replaced by 0. Given the input, the decoder predicts the sequence in the next 15 days ${\mathop {\widehat {\textit{PR}}}\nolimits_{{\mathop t\nolimits_0} + 1,{\mathop t\nolimits_0} + 15}^{{\mathop t\nolimits_0} }}$. Therefore, the sizes of the input data and time encoding are $43 \times \mathop n\nolimits_{new}$ and $43 \times 4$ respectively. A generative style decoder is adopted to acquire long sequence output with only one forward step needed, simultaneously avoiding cumulative error spreading during the inference.

\subsection{Confidence Estimation Method}

Combining the prediction results from different time can offset the contingency of single prediction result. This paper designs a confidence estimation method to represent the effectiveness of a prediction sequence as the combining weight. 

$P\left(e_i \mid B\right)$ is the condition probability that the $i^{th}$ result in the prediction sequence is correct given the condition $B$ which means the accuracy of partial results is observed. $P\left(e_i \mid B\right)$ can be derived as

\begin{equation}
P\left(e_i \mid B\right)=\frac{P\left(e_i\right) P\left(B \mid e_{i}\right)}{\sum_{j=1}^{n} P\left(e_{j}\right) P\left(B \mid e_{j}\right)}
\end{equation}
To facilitate the update of effectiveness, we use beta distribution to represent confidence of the prediction sequence according to the accuracy of historical prediction results, which is defined as

\begin{equation}
\begin{aligned}
f(x ; m, n)&=\frac{x^{m-1}(1-x)^{n-1}}{\int_{0}^{1} u^{m-1}(1-u)^{n-1} d u}\\
&=\frac{\Gamma(m+n)}{\Gamma(m) \Gamma(n)} x^{m-1}(1-x)^{n-1}\\
&=\frac{1}{B(m, n)} x^{m-1}(1-x)^{n-1}\\
\end{aligned}
\end{equation}
where $ m>0 \text { and } n>0 $ represent the number of correct prediction results and incorrect prediction results, respectively. In this paper, the prediction result is judged to be correct, when its root mean square error is less than ${\mu }$. $\Gamma(x)$ is the Gamma function, which could be written as

\begin{equation}
\Gamma(x)=\int_{0}^{+\infty} t^{x-1} e^{-t} \mathrm{~d} t(x>0)
\end{equation}


When the prediction model is initialized without prior knowledge, the confidence of the prediction sequence is expressed as a uniform distribution on $(0,1)$:

\begin{equation}
P(x)=\operatorname{uni}(0,1)=\operatorname{Beta}(1,1)
\end{equation}

Suppose that $t = {t_0} - d\left( {1 \leqslant d \leqslant 5} \right)$ and the prediction sequence at time $t$ is ${\mathop {\widehat {\textit{PR}}}\nolimits_{{\mathop {t+1}},{\mathop {t+15}}}^{{\mathop t}}}$. As shown in Fig. \ref{fig_4} b, the confidence estimation method can evaluate the effectiveness of prediction sequence at time $t$ by comparing ${\mathop {\widehat {\textit{PR}}}\nolimits_{{\mathop {t+1}},{\mathop t\nolimits_0}}^{{\mathop t}}}$ with ${\mathop { \textit{PR}}\nolimits_{{\mathop {t+1}},{\mathop t\nolimits_0}}}$. It is assumed that $m$ correct prediction results and $n$ incorrect prediction results are identified after comparison. The estimated confidence of the prediction sequence is defined as
\begin{equation}
\begin{aligned}
  Con_t={\log _{10}}(d + 1)E[Beta(m,n)] 
   =\frac{{m{{\log }_{10}}(d + 1)}}{{m + n}} 
\end{aligned} 
\label{eq_18}
\end{equation}
Finally, the confidence is normalized as
\begin{equation}
{\lambda _t} = \frac{{C{on_t}}}{{\sum\limits_{i = \mathop t\nolimits_0  - 5}^{\mathop t\nolimits_0  - 1} C {on_i}}}
\label{eq_19}
\end{equation}

\subsection{Process of Conformer}\label{cp}


The proposed Conformer algorithm is shown in Algorithm \ref{alg_1}. Firstly, the enhanced Transformer-based renewable energy prediction model predicts the sequence in the next 15 days ${\mathop {\widehat {\textit{PR}}}\nolimits_{{\mathop t\nolimits_0} + 1,{\mathop t\nolimits_0} + 15}^{{\mathop t\nolimits_0} }}$ on the basis of the real data in the past 56 days ${\mathop { \textit{PR}}\nolimits_{{\mathop t\nolimits_0} -55,{\mathop t\nolimits_0}}}$. And the prediction results are stored to the storage pool $D_s$. If ${\mathop t\nolimits_0}<5$, ${\mathop {\widehat {\textit{PR}}}\nolimits_{{\mathop t\nolimits_0} + 1,{\mathop t\nolimits_0} + 15}^{{\mathop t\nolimits_0} }}$ is directly intercepted as ${\mathop {\widehat {\textit{PR}}}\nolimits_{{\mathop t\nolimits_0} + 1,{\mathop t\nolimits_0} + 10}}$. If ${t_0} \geqslant5$, the past 5 days' prediction sequences are taken out from the storage pool $D_s$, and the confidence of the sequences is calculated via \eqref{eq_18} and \eqref{eq_19}. Finally, the predicted active output results from ${\mathop t\nolimits_0} + 1$ to ${\mathop t\nolimits_0}+10$ in the past 5 days' prediction sequences are multiplied by the corresponding estimated confidence respectively and then accumulated to obtain the effective prediction ${\mathop {\widehat {\textit{PR}}}\nolimits_{{\mathop t\nolimits_0} + 1,{\mathop t\nolimits_0} + 10}}$. 

In the prediction process of Conformer, this paper combines the enhanced Transformer prediction results from different time to get the final prediction, thereby reducing the prediction error. The final prediction is delivered to RLpatching to assist the power grid dispatching and abate the impact of new energy uncertainty.

\begin{algorithm}[htbp]\label{alg_1}
            \caption{Conformer Prediction Algorithm}
            
            \KwIn{Current time ${\mathop t\nolimits_0}$, the enhanced Transformer-based renewable energy prediction model \emph{REPM} and the storage pool $D_s$}
            \KwOut{Prediction results ${\mathop {\widehat {\textit{PR}}}\nolimits_{{\mathop t\nolimits_0} + 1,{\mathop t\nolimits_0} + 10}}$}
            \LinesNumbered
            
            Predict the ${\mathop {\widehat {\textit{PR}}}\nolimits_{{\mathop t\nolimits_0} + 1,{\mathop t\nolimits_0} + 15}^{{\mathop t\nolimits_0} }}$ by \emph{REPM}\;
            Store ${\mathop {\widehat {\textit{PR}}}\nolimits_{{\mathop t\nolimits_0} + 1,{\mathop t\nolimits_0} + 15}^{{\mathop t\nolimits_0} }}$ to $D_s$\;
            \eIf{${\mathop t\nolimits_0}<5$}{
           Intercept ${\mathop {\widehat {\textit{PR}}}\nolimits_{{\mathop t\nolimits_0} + 1,{\mathop t\nolimits_0} + 10}}$ in ${\mathop {\widehat {\textit{PR}}}\nolimits_{{\mathop t\nolimits_0} + 1,{\mathop t\nolimits_0} + 15}^{{\mathop t\nolimits_0} }}$\;
           }
           {
           Get ${\mathop{\widehat{\textit{PR}}}\nolimits_{i + 1,i + 15}^i}$  $\left( {i = {\mathop t\nolimits_0} - 5,...,{\mathop t\nolimits_0} - 1} \right)$ from $D_s$\;
           Calculate the estimated confidence ${\lambda _i}$ via \eqref{eq_18}, \eqref{eq_19}\;
           Calculate the final prediction results ${\widehat {\textit{PR}}_{{\mathop t\nolimits_0} + 1,{\mathop t\nolimits_0} + 10}} = \sum\limits_{i = {\mathop t\nolimits_0} - 5}^{{\mathop t\nolimits_0} - 1} {\mathop \lambda \nolimits_i \mathop {\widehat{\textit{PR}}}\nolimits_{{\mathop t\nolimits_0} + 1,{\mathop t\nolimits_0} + 10}^i } $\;
           }
           
          \Return ${\mathop {\widehat {\textit{PR}}}\nolimits_{{\mathop t\nolimits_0} + 1,{\mathop t\nolimits_0} + 10}}$\;
    
    \end{algorithm}

\section{Reinforcement Learning-based Dispatching for the Hybrid Power Grid}\label{dispatching}
This section mainly introduces RL-based multi-objective dispatching framework. Firstly, the optimization objectives and constraints are illustrated in Section \ref{OO}. DDPG-based power flow optimization, the core algorithm of RLpatching, is described in Section \ref{DDPG}. Next, Section \ref{evaluation} presents the dispatching necessity evaluation mechanism. Finally, Section \ref{process} introduces the decision-making and training process of RLpatching with the assistance of Conformer.


\subsection{Optimization Objectives and Constraint Conditions}\label{OO}
The optimization objectives consist of security, economic and environmental indicators. The security objective considers branch overflow ${\mathop S\nolimits_b^t}$, reactive power output overrun ${\mathop S\nolimits_r^t}$ and voltage overrun ${\mathop S\nolimits_v^t}$. The economic objective aims at minimizing the cumulative cost of the power system ${\mathop C\nolimits^t}$, which includes the operation cost and startup cost of the generators. And the environmental objective targets to maximize URRE $\mathop R\nolimits^t$. 
The formula of the optimization objectives is
\begin{gather}
\begin{aligned}
&\max \{ \sum\limits_{t = 1}^{\mathop N\nolimits_{steps} } {(\zeta (  \mathop S\nolimits_r^t ) + \mathop S\nolimits_b^t   + \zeta (  \mathop S\nolimits_v^t ) + \zeta ( - \mathop C\nolimits^t ) + \mathop \omega \nolimits_R \mathop R\nolimits^t )\} }\label{eq_8}\\
&\emph{in which,}
\end{aligned}\\
\mathop S\nolimits_r^t  = \sum\limits_{i = 1}^{\mathop n\nolimits_{gen} }{\left\{{\begin{array}{*{20}{l}}{{\mathop u\nolimits_{i}^t }(1 - \frac{{{\rm{ }}{q_i^t}}}{{{\rm{ }}q_i^{\max }}})}&{{q_i^t} > {\rm{ }}q_i^{\max }}\\{{\mathop u\nolimits_{i}^t }(\frac{{{\rm{ }}{q_i^t}}}{{{\rm{ }}q_i^{\min }}} - 1)}&{{q_i^t} < {\rm{}}q_i^{\min}}\\0&{else}\end{array}} \right.}\tag{\ref{eq_8}{a}}\\
\mathop S\nolimits_b^t  = 1 - \frac{1}{{{n_{branch}}}}\sum\limits_{j = 1}^{\mathop n\nolimits_{branch} } {\min (\frac{{{\mathop I\nolimits_{j}^t}}}{{{T_j}}},1)}\tag{\ref{eq_8}{b}}\\
\mathop S\nolimits_v^t  = \sum\limits_{k = 1}^{\mathop n\nolimits_{bus} } {\left\{ {\begin{array}{*{20}{l}}{1 - \frac{{{\rm{ }}{v_k^t}}}{{{\rm{ }}v_k^{\max }}}}&{{v_k^t} > {\rm{ }}v_i^{\max }}\\{\frac{{{\rm{ }}{v_k^t}}}{{{\rm{ }}v_k^{\min }}} - 1}&{{v_k^t} < {\rm{ }}v_k^{\min }}\\0&{else}\end{array}} \right.}\tag{\ref{eq_8}{c}}\\
{\mathop C\nolimits^t} =  {\sum\limits_{i = 1}^{\mathop n\nolimits_{gen} } {(\mathop u\nolimits_i^t (\mathop a\nolimits_i \mathop {(\mathop P\nolimits_i^t )}\nolimits^2  + \mathop b\nolimits_i \mathop P\nolimits_i^t  + \mathop c\nolimits_i )  +  \mathop u\nolimits_i^t (1 - \mathop u\nolimits_i^{t - 1} )\mathop c\nolimits_i^{start} )} } \tag{\ref{eq_8}{d}}\\
\mathop R\nolimits^t  = \frac{{\sum\limits_{i = 1}^{{n_{new}}} {\mathop P\nolimits_i^t } }}{{\sum\limits_{i = 1}^{{n_{new}}} {\mathop P\nolimits_i^{\max } } }}\tag{\ref{eq_8}{e}}
\end{gather}


Here, ${\mathop N\nolimits_{steps}}$ is the maximum number of steps in which the dispatching strategy can make the simulator run safely under the constraints. $\mathop \omega \nolimits_R$ is the weight coefficient greater than one to improve URRE.
${\mathop n\nolimits_{gen}}$, ${\mathop n\nolimits_{branch} }$, ${\mathop n\nolimits_{bus} }$ and ${\mathop n\nolimits_{new}}$ represent the number of generators, branches, busbars and renewable energy generators, respectively. ${{\mathop u\nolimits_{i}^t }}$, ${\mathop q\nolimits_i^t}$ and ${{\mathop P\nolimits_{i}^t }}$ are the on/off status, reactive power output and active output of generator $i$ in period $t$, respectively. ${\mathop q\nolimits_i^{\max}}$ and ${\mathop q\nolimits_i^{\min}}$ represent the maximum and minimum reactive power output of generator $i$, and ${\mathop P\nolimits_i^{\max}}$ is the maximum active output of generator $i$. ${\mathop I\nolimits_{j}^t }$ and ${\mathop T\nolimits_j }$ separately represent the current and thermal limits of branch $j$. ${\mathop v\nolimits_k^t}$ is the voltage of busbar $k$ in period $t$, ${\mathop v\nolimits_k^{\max}}$ and ${\mathop v\nolimits_k^{\min}}$ represent the maximum and minimum voltage of busbar $k$, respectively. ${\mathop a\nolimits_i}$, ${\mathop b\nolimits_i}$ and ${\mathop c\nolimits_i}$ are operation cost factors of generator $i$. ${\mathop c\nolimits_i^{start}}$, the factor of startup cost, is a fixed value for generator $i$.
In addition, we use a normalization function ${\zeta(x) = \mathop e\nolimits^x-1}$ to limit the ranges of ${\mathop S\nolimits_r^t}$, ${\mathop S\nolimits_v^t}$ and ${-\mathop C\nolimits^t}$ to $\left( { - 1 ,0} \right]$.

In this paper, three constraint conditions are set to limit optimization.
\subsubsection{Voltage constraints}
The actual value of the voltage of any generator should not be greater than the upper limit of the voltage, nor less than the lower limit. Otherwise, the security objective will be affected negatively.
\subsubsection{Power balance constraints}
The total power generation should cover the total power demand. Hence,
\begin{equation}
   {\sum\limits_m^{\mathop n\nolimits_t } {\mathop P\nolimits_{Tm}^t }  + \sum\limits_n^{\mathop n\nolimits_{new} } {\mathop P\nolimits_{Rn}^t }   = \sum\limits_j^{\mathop n\nolimits_l } {\mathop P\nolimits_{Lj}^t }}
\end{equation}
where ${\mathop n\nolimits_t}$ is the number of thermal generators, and ${\mathop n\nolimits_l}$ is the number of loads.
\subsubsection{Ramp rate constraints}
The active power adjustment value of each thermal power generator between any two continuous time steps must be smaller than the generator maximum adjustment value. Therefore, for generator $i$ in period $t$, the ramp rate constraint is defined as
\begin{equation}
    {|\mathop P\nolimits_i^t  - \mathop P\nolimits_i^{t + \Delta t} | < \mathop {ramp\_rate \times P}\nolimits_i^{max}}
\end{equation}

\subsection{DDPG-based Power Flow Optimization}\label{DDPG}
\begin{figure}[h]
\centering
\centerline{\includegraphics[width=0.9\columnwidth]{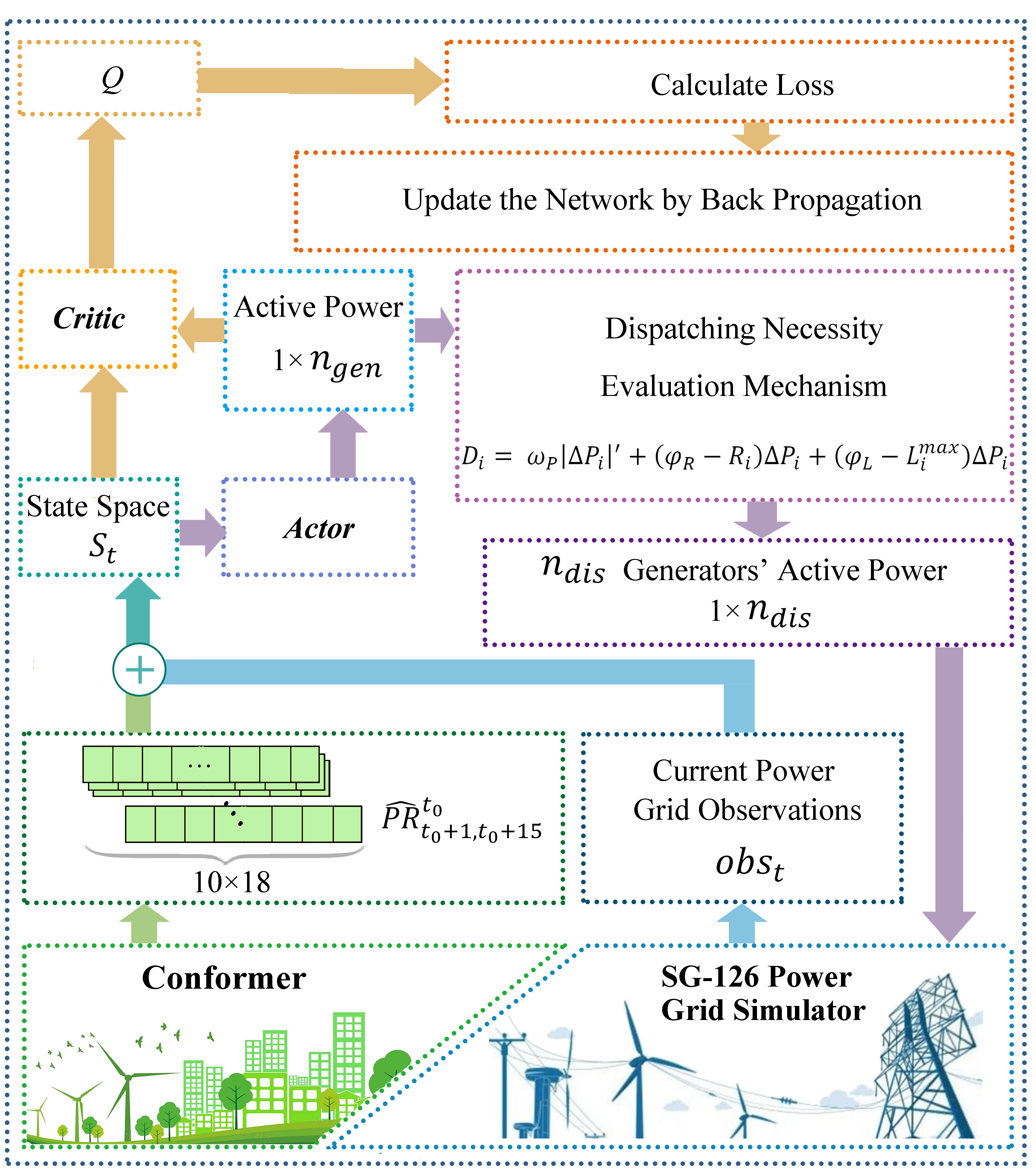}}
\caption{Architecture of RLpatching}
\label{fig_2}
\end{figure}

RLpatching adopting DDPG determines the active power output of all generators for the next day according to the current power grid operation state and long-term prediction of renewable energy from Conformer. It includes ${actor}$ and ${critic}$, which are used for decision-making and scoring, respectively, as shown in Fig. \ref{fig_2}.

In RLpatching, the input of ${actor}$ includes the current power grid observations $obs_t$ and the final prediction ${\mathop {\widehat {\textit{PR}}}\nolimits_{{\mathop t} + 1,{\mathop t} + 10}}$ output by Conformer, which are marked as the state space $\mathop s\nolimits_t$. $obs_t$ consists of 13 selected groups of current power grid parameters to represent the operation state, including reactive power of all generators, voltage, active power and reactive power at the starts and ends of branches, load ratio and current of branches, active power and reactive power of the loads and the grid loss. ${actor}$ outputs active power of all generators for the next day  $\mathop P^t$. ${critic}$ takes $\mathop s\nolimits_t$, $\mathop P^t$ as its input and outputs the ${value}$ ${Q}$. 


Since the grid operation control is modeled as a time-continuous decision-making process, the reward function is defined as the optimization objectives at each time step. The reward function at time $t$ is 
\begin{equation}
{\mathop r\nolimits_t  = \mathop S\nolimits_b^t  + \zeta ( \mathop S\nolimits_r^t ) + \zeta ( \mathop S\nolimits_v^t ) + \zeta ( - \mathop C\nolimits^t ) + \mathop \omega \nolimits_R \mathop R\nolimits^t}
\end{equation}

The ${actor}$ and ${critic}$ parameterized by ${\mathop \theta \nolimits^\mu}$, ${\mathop \theta \nolimits^Q }$ are used to represent the deterministic policy ${\mathop P^t  = \mu (\mathop s\nolimits_t \left| {\mathop \theta \nolimits^\mu  } \right.)}$ and the critic function ${Q(\mathop s\nolimits_t ,\mathop P^t \left| {\mathop \theta \nolimits^Q } \right.)}$. ${\mathop \theta \nolimits^\mu}$ and ${\mathop \theta \nolimits^Q }$ are optimized by stochastic gradient method. The loss functions of $actor$ and $critic$ are
\begin{equation}
    J({\theta ^\mu }) = \frac{1}{N}\sum\limits_t {{{\left. {Q({s_t},{P^t}\left| {{\theta ^Q}} \right.)} \right|}_{{P^t} = \mu \left( {{s_t}} \right)}}\mu \left( {{s_t}\mid {\theta ^u}} \right)}
\end{equation}

\begin{equation}
    J({\theta ^Q}) = \frac{1}{N}\sum\limits_t {\mathop {[{r_t} + \gamma Q'({s_{t + 1}},{{\mu '}_{t + 1}}\left| {{\theta ^{Q'}}} \right.) - Q({s_t},{P^t}\left| {{\theta ^Q}} \right.)]}\nolimits^2 }
\end{equation}


\subsection{Dispatching Necessity Evaluation Mechanism}\label{evaluation}
In power grid dispatching, adjusting numerous generators would decrease the stability sharply. To deal with this problem, this paper designs a dispatching necessity evaluation mechanism to quantify the necessity of each generator and select $\mathop n\nolimits_{dis}$ generators to dispatch. For generator $i$, three factors are considered to evaluate its necessity, including active power adjustment value for the next day $\mathop {\Delta P}\nolimits_i$, utilization rate ${R_i} = \frac{{\mathop P\nolimits_i^t }}{{\mathop P\nolimits_i^{\max } }}$, and maximum load rate of all branches around it $\mathop {L}\nolimits_i^{max}$. $\left| {\mathop {\Delta P}\nolimits_i } \right|$ is normalized as ${\mathop {\left| {\mathop {\Delta P}\nolimits_i } \right|}\nolimits^{'}}$ by the min-max method.
We select $\mathop n\nolimits_{dis}$ generators with the highest necessity to dispatch, and the active output of other generators is consistent with that of previous time step. The necessity formula of dispatching is defined as
\begin{equation}
    \mathop D\nolimits_i  = \mathop {\mathop \omega \nolimits_P \left| {\mathop {\Delta P}\nolimits_i } \right|}\nolimits^{'}  + ( {\mathop \varphi \nolimits_R}-\mathop R\nolimits_i)\mathop {\Delta P}\nolimits_i  + ( {\mathop \varphi \nolimits_L}-\mathop L\nolimits_i^{\max })\mathop {\Delta P}\nolimits_i
\label{eq_12}
\end{equation}

Equation \eqref{eq_12} is a polynomial composed of three parts. $\mathop {\mathop \omega \nolimits_P \left| {\mathop {\Delta P}\nolimits_i } \right|}\nolimits^{'}$ measures the adjustment value, with which the dispatching necessity is positively correlated. $( {\mathop \varphi \nolimits_R}-\mathop R\nolimits_i)\mathop {\Delta P}\nolimits_i$ is designed to control the utilization rate of 
generator $i$ within a reasonable range. When utilization rate of generator is less than ${\mathop \varphi \nolimits_R}$, the necessity of adjustment increases with the increase of $\mathop {\Delta P}\nolimits_i$. When utilization rate of generator is greater than ${\mathop \varphi \nolimits_R}$, the increase of active power of generator $i$ is restrained. $( {\mathop \varphi \nolimits_L}-\mathop L\nolimits_i^{\max })\mathop {\Delta P}\nolimits_i$ maintains the load rate of branches around generator $i$, featuring the same characteristics as $( {\mathop \varphi \nolimits_R}-\mathop R\nolimits_i)\mathop {\Delta P}\nolimits_i$.


\subsection{RLpatching Decision-making and Training Process}\label{process}
The decision-making and training process of the proposed Conformer-RLpatching for the hybrid power grid is shown in Algorithm \ref{alg_2}. At the beginning of each epoch,  $obs_{\mathop t\nolimits_0}$ is obtained from the initialized simulator, together with which the prediction from Conformer initializes the state space $\mathop s\nolimits_{{\mathop t\nolimits_0}}$. 
In each step of dispatching, active power output of all generators for the next step is determined by the $actor$, the dispatching necessity evaluation mechanism selects $\mathop n\nolimits_{dis}$ generators to dispatch, and then the simulator executes the dispatching strategy. Next, $obs_{{\mathop t\nolimits_0} + 1}$ and ${\mathop {\widehat {\textit{PR}}}\nolimits_{{\mathop t\nolimits_0} + 2,{\mathop t\nolimits_0} + 11}}$ consists $\mathop s\nolimits_{{\mathop t\nolimits_0} + 1}$.
Afterwards $(\mathop s\nolimits_{\mathop t\nolimits_0}, \mathop P^{\mathop t\nolimits_0}, \mathop r\nolimits_{\mathop t\nolimits_0} , \mathop s\nolimits_{{\mathop t\nolimits_0}+1})$ is stored to $D_e$. Finally, samples are selected from $D_e$ to update all networks. This epoch will end when the power flow calculation of the simulator fails to converge or the operation of the grid does not meet the constraints mentioned in Section \ref{OO}.


\begin{algorithm}[h]\label{alg_2}
            \caption{RLpatching Decision-making and Online Training Algorithm}
            
            \LinesNumbered

            Initialize $actor$, $critic$ and Conformer\;
           \For{epoch=1:R}{
           Initialize experience pool $D_e$ and storage pool $D_s$\;
           ${\mathop t\nolimits_0}=0$\;
           Initialize the SG-126 power grid simulator and  Obtain $obs_{\mathop t\nolimits_0}$\;
           Predict ${\mathop {\widehat {\textit{PR}}}\nolimits_{{\mathop t\nolimits_0} + 1,{\mathop t\nolimits_0} + 10}}$ by Alg. \ref{alg_1}\;
           $[{\mathop {obs}\nolimits_{\mathop t\nolimits_0}},{\mathop {\widehat {\textit{PR}}}\nolimits_{{\mathop t\nolimits_0} + 1,{\mathop t\nolimits_0} + 10}}]\rightarrow {\mathop s\nolimits_{\mathop t\nolimits_0}}$\;
           \While{True}{
           Input ${\mathop s\nolimits_{\mathop t\nolimits_0}}$ to $actor$ and get $P^{\mathop t\nolimits_0}$\;
           Select $\mathop n\nolimits_{dis}$ generators to dispatch via \eqref{eq_12}\;
           Observe ${\mathop r\nolimits_{\mathop t\nolimits_0}}$, $Done$, ${\mathop {obs}\nolimits_{{\mathop t\nolimits_0}+1}}$ and ${{\textit{PR}}_{{\mathop t\nolimits_0} + 1}}$\ from the simulator\;
           Predict ${\mathop {\widehat {\textit{PR}}}\nolimits_{{\mathop t\nolimits_0} + 2,{\mathop t\nolimits_0} + 11}}$ by Algorithm \ref{alg_1}\;
           $[{\mathop {obs}\nolimits_{{\mathop t\nolimits_0}+1}},{\mathop {\widehat {\textit{PR}}}\nolimits_{{\mathop t\nolimits_0} + 2,{\mathop t\nolimits_0} + 11}}]\rightarrow {\mathop s\nolimits_{{\mathop t\nolimits_0}+1}}$\;
           Store $(\mathop s\nolimits_{\mathop t\nolimits_0}, \mathop P^{\mathop t\nolimits_0}, \mathop r\nolimits_{\mathop t\nolimits_0} , \mathop s\nolimits_{{\mathop t\nolimits_0}+1})$ to $D_e$\;
           Select samples from $D_e$ and update networks\;
           \If{$Done$}{
           break\;
           }
           ${\mathop t\nolimits_0}={\mathop t\nolimits_0}+1$\;
           $\mathop s\nolimits_{{\mathop t\nolimits_0}}=\mathop s\nolimits_{{\mathop t\nolimits_0}+1}$\;
           }
           }
    
\end{algorithm}

\section{Simulation and Results}\label{results}

\subsection{The Simulator and Settings}\label{SCM}

Conformer-RLpatching is evaluated on SG-126 power grid simulator. The simulator includes 54 generators and 117 branches, which conforms to the characteristics and operation mode of provincial grid. 
The simulation parameters and generator parameters are illustrated in TABLE \ref{table_par} and TABLE \ref{table_grid}, respectively.

\begin{table}[H]
\caption{Simulation parameters}
\centering
\begin{tabular}{|c|c|c|c|}
\hline
\textbf{Parameters} & \textbf{Value} & \textbf{Parameters} & \textbf{Values} \\ \hline
       Renewable energy units&            18    &  Thermal power units&            36        \\ \hline
       Branches              &            185   &   Loads                 &             91   \\ \hline
       Transformers          &            9     &   Ramp rate         &             0.05     \\ \hline
       ${\mathop \omega \nolimits_R}$&   2      &   ${\mu }$           &   5  \\ \hline
\end{tabular}
\label{table_par}
\end{table}

\begin{table}[H]
\centering
\caption{Generator Parameters}
\label{table_grid}
 \scalebox{0.9}[1]{
\begin{tabular}{ccccc}
\hline
\multirow{2}{*}{\textbf{Parameters}} & \multicolumn{4}{c}{\textbf{Units}}                                                                                                                                                                                                       \\ \cline{2-5} 
                                     & \textbf{U1-U18}                                            & \textbf{U19-U30}                                        & \textbf{U30-U40}                                        & \textbf{U40-U54}                                        \\ \hline
\textbf{Type}                        & \begin{tabular}[c]{@{}c@{}}Renewable\\ Energy\end{tabular} & \begin{tabular}[c]{@{}c@{}}Thermal\\ Power\end{tabular} & \begin{tabular}[c]{@{}c@{}}Thermal\\ Power\end{tabular} & \begin{tabular}[c]{@{}c@{}}Thermal\\ Power\end{tabular} \\
\bm{${\mathop P\nolimits_i^{max}/MW}$}                    & -                                                         & 110                                                     & 128                                                     & 140                                                     \\
\bm{${\mathop P\nolimits_i^{min}/MW}$}                      & 0                                                          & 15                                                      & 25                                                      & 28                                                      \\
\bm{${\mathop v\nolimits_i^{max}/Mm^3}$}                     & 105                                                        & 105                                                     & 105                                                     & 105                                                     \\
\bm{${\mathop v\nolimits_i^{min}/Mm^3}$}                     & 95                                                         & 95                                                      & 95                                                      & 95                                                      \\
\bm{${\mathop a\nolimits_i}$}                           & 0.0696                                                     & 0.0285                                                  & 0.0109                                                  & 0.0097                                                  \\
\bm{${\mathop b\nolimits_i}$}                           & 26.2438                                                    & 17.82                                                   & 22.9423                                                 & 12.8875                                                 \\
\bm{${\mathop c\nolimits_i}$}                          & 31.67                                                      & 10.15                                                   & 32.96                                                   & 58.81                                                   \\
\bm{${\mathop c\nolimits_i^{start}}$}                          & 80                                                         & 100                                                     & 200                                                     & 880                                                     \\ \hline
\end{tabular}}
\end{table}

\begin{table*}[]
\caption{Prediction RMSE}
\label{RMSE}
\setlength\tabcolsep{4pt}
\begin{center}
\begin{tabular}{ccccccccccccc}
\toprule
Input data length  & \multicolumn{4}{c}{48}                                                                            & \multicolumn{4}{c}{56}                                                                            & \multicolumn{4}{c}{64}                                                                            \\ \hline
Output data length & \multicolumn{1}{c}{15} & \multicolumn{1}{c}{20} & \multicolumn{1}{c}{25} & \multicolumn{1}{c}{30} & \multicolumn{1}{c}{15} & \multicolumn{1}{c}{20} & \multicolumn{1}{c}{25} & \multicolumn{1}{c}{30} & \multicolumn{1}{c}{15} & \multicolumn{1}{c}{20} & \multicolumn{1}{c}{25} & \multicolumn{1}{c}{30} \\ \hline
Conformer          & \textbf{5.0648}                & \textbf{5.3919}             & \textbf{5.8599}           &  \textbf{6.2192}                      &   \textbf{4.9995}          &      \textbf{5.2718}        &   \textbf{5.9732}            & \textbf{6.1780}                    & \textbf{5.0076}        &     \textbf{5.3985}     & \textbf{6.0351}       &  \textbf{6.1866}                     \\
Informer               & 5.9370                  & 6.7637                 & 7.2799                 & 7.8346                 & 5.9677                 & 6.6830                  & 7.3198                 & 7.7981                 & 6.0773                 & 6.7428                 & 7.3137                 & 7.7993                 \\
LSTMa              & 10.9698                & 11.3034                & 12.1275                & 12.2250                 & 10.8640                & 12.7792                & 13.0704                & 12.5671                & 11.1334                & 12.8136                & 13.0573                & 12.9230                 \\
Prophet            & 8.9671                 & 9.7796                 & 10.8414                & 11.8916                & 9.4211                 & 10.3752                & 11.2745                & 12.3669                & 9.5111                 & 10.3964                 & 11.2522                & 12.5742                \\
CNN                & 6.6045                 & 7.5408                 & 8.0617                 & 8.5013                 & 6.7521                 & 7.4893                 & 8.0038                 & 8.5720                  & 6.8096                 & 7.5571                 & 7.9928                 & 8.6852                 \\
CNN-LSTM           & 7.8669                 & 8.6973                 & 9.2094                 & 9.5646                 & 8.2419                 & 9.3553                 & 9.6920                  & 9.7102                 & 8.3122                  & 9.4028                 & 9.4484                 & 9.9086                 \\ \bottomrule
\end{tabular}
\end{center}
\end{table*}

\begin{figure*}[t]
    \centering
    \centerline{\includegraphics[width=0.9\textwidth,height=4.7cm]{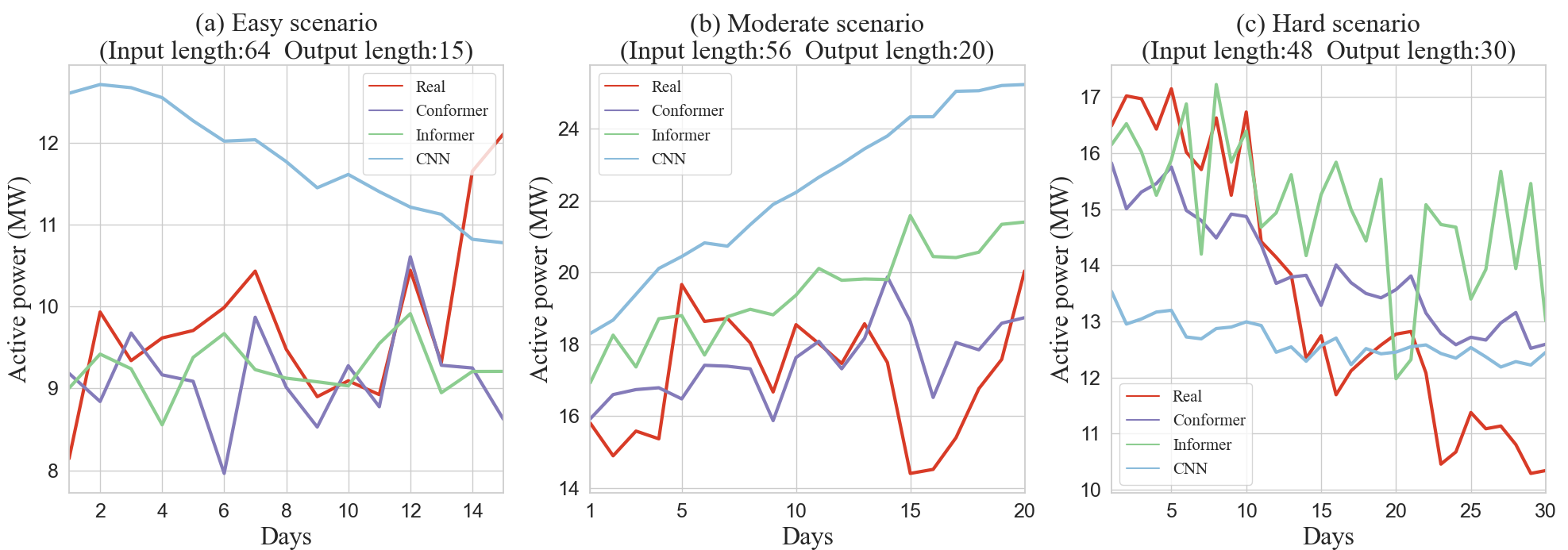}}
    \caption{Prediction Results}
    \label{pre_result}
\end{figure*}

\subsection{Comparison of Renewable Energy Prediction Models}\label{prediction result}

 This subsection compares the predictive effectiveness of Conformer with that of state-of-the-art methods, including Informer \cite{d9}, LSTMa \cite{e1}, Prophet \cite{e2}, convolutional neural network (CNN), and CNN-LSTM \cite{d8}. The data set used in the experiment is the maximum active power output of 18 renewable energy units in 106820 days, which is provided by China Electric Power Research Institute. This paper selects 90$\%$ of the data as the training set and 10$\%$ as the testing set. The root mean square error (RMSE) is used to evaluate the effect of the models:
\begin{equation}
    RMSE = \sqrt {\frac{1}{{{t_{pre}}{n_{new}}}}\sum\limits_t^{{t_{pre}}} {\sum\limits_i^{{n_{new}}} {{{(P_i^t - \hat P_i^t)}^2}} } } 
\end{equation}

We set up a series of comparative experiments, w.r.t variable input length and output length, to predict the long-term maximum active output of all renewable energy generators. In these experiments, historical data of the past 48 days, 56 days, 64 days, 72 days, 80 days, 88 days, and 96 days are fed into models separately, and the length of predicted values ${t_{pre}}$ is set as 15, 20, 25 and 30. We only demonstrate the crutial results in TABLE \ref{RMSE}, the complete results are exhibited in APPENDIX \ref{App}. 

Conformer achieves much better performance than all compared methods in the overall experiments, which demonstrates its superiority. Specifically, the average RMSE of Conformer in 12 groups of comparative experiments is 19.07$\%$ lower than that of Informer with the second-best performance, which means Conformer has less prediction error. The results show that the confidence estimation method offsetting the contingency of a single prediction result facilitates the predictive effectiveness. In addition, the prediction accuracy of Conformer is 39.65$\%$ higher than other algorithms on average.

When the input length is fixed, all methods show a consistent trend, that is, the gap between the prediction and real data widens slightly as the output length increases. For example, when the input length is 56 and the output length rises from 15 to 30, the RMSE of Conformer and CNN increases by 23.57$\%$ and 26.95$\%$, respectively.

Given the fixed output length, the prediction accuracy of all methods tends to increase with the input length rising, and reaches a peak as the input length is approximately 56. The phenomenon illustrates that more historical data endows the model with better predictive ability, while excessive input will bring more interference of invalid data to the model, thus reducing the prediction accuracy. When the output length is fixed at 20, despite the predictive error of Conformer increasing by 2.40$\%$ as the input length rises from 56 to 64, Conformer still maintains its superiority with the RMSE 19.94$\%$, 28.56$\%$ lower than that of Informer and CNN, respectively. 

This paper further observes the robustness of Conformer in different conditions of variable input and output lengths compared with Informer and CNN. Short-term prediction with more historical data makes it much easier for models. Therefore, three scenarios of variable difficulty are set as shown Fig. \ref{pre_result}. It is clear that CNN has the worst performance in all scenarios. As depicted in Fig. \ref{pre_result} (a), both Conformer and Informer are able to describe the correct trend of real data, specially the prediction from day 9 to day 13 output by Conformer achieves surprisingly little error. In the moderate scenario Fig. \ref{pre_result} (b), Conformer performs more stably than Informer, and achieves more accurate long-term prediction. Conformer significantly outperforms Informer in the hard scenario Fig. \ref{pre_result} (c), where Informer deviates real data as output length rises, whereas Conformer describes the correct trend of real data accurately. In summary, Conformer still achieves the best performance despite the prediction accuracy decreasing as the difficulty rises, which proves its advantageous robustness.


\begin{table*}[t]
\caption{Simulation Results}
\label{SiR}
\begin{center}
\begin{threeparttable}

\begin{tabular}{lcccccccc}
\toprule
                          & \bm{${\mathop \varphi \nolimits_R}$}   & \bm{${\mathop \varphi \nolimits_L}$}   & \bm{$\mathop n\nolimits_{dis}$}   & \textbf{Steps} & \begin{tabular}[c]{@{}c@{}}\textbf{Security} \\ \textbf{Score}\end{tabular} & \begin{tabular}[c]{@{}c@{}}\textbf{Average Cost} \\ \textbf{(Thousand RMB Yuan)}\end{tabular} & \begin{tabular}[c]{@{}c@{}}\textbf{Average Renewable} \\ \textbf{Energy Utilization Rate}\end{tabular} & \begin{tabular}[c]{@{}c@{}}\textbf{Total}\\ \textbf{Reward}\end{tabular} \\ \hline
Conformer-RLpatching I    & 0.8 & 0.8 & 40 & \textbf{426}   & \textbf{261.145}                                                   & 65251.618                                               & 81.579$\%$                                                                           & \textbf{527.318}                                                \\
Conformer-RLpatching II   & 0.8 & 0.8 & 40 & 309      & 202.073                        & \textbf{56499.215}                                                        &   73.826$\%$                                                          & 346.117                                                       \\ \hline
Conformer-RLpatching III  & 0.7 & 0.8 & 40 & 417   & 250.544                                                   & 65552.791                                               & 79.798$\%$                                                                          & 498.990                                                 \\
Conformer-RLpatching IV   & 0.9 & 0.8 & 40 & 415   & 254.921                                                   & 65399.495                                               & 77.909$\%$                                                                           & 485.560                                                 \\
Conformer-RLpatching V    & 0.8 & 0.7 & 40 & 412   & 252.695                                                   & 65651.850                                                & 81.822$\%$                                                                           & 510.402                                                \\
Conformer-RLpatching VI   & 0.8 & 0.9 & 40 & 395   & 241.744                                                   & 66163.359                                               & 82.754$\%$                                                                           & 497.888                                                \\
Conformer-RLpatching VII  & 0.8 & 0.8 & 35 & 408   & 252.815                                                   & 65538.547                                               & 80.055$\%$                                                                           & 498.064                                                \\
Conformer-RLpatching VIII & 0.8 & 0.8 & 45 & 417   & 245.887                                                   & 66081.338                                               & 82.638$\%$                                                                           & 513.476                                                \\ \hline
\multicolumn{1}{c}{A+B}   & -   & -   & -  & 411   & 250.824                                                   & 65512.642                                               & 79.198$\%$                                                                           & 490.008                                                \\
\multicolumn{1}{c}{B+C}   & 0.8 & 0.8 & 40 & 206   & 129.021                                                   & 68862.785                                               & 78.333$\%$                                                                           & 241.563                                                \\ \hline
\multicolumn{1}{c}{DDPG}  & -   & -   & -  & 325   & 207.656                                                   & 63973.590                                                & 55.871$\%$                                                                           & 243.303                                                \\
\multicolumn{1}{c}{DCR-TD3}   & -   & -   & -  & 250   & 159.899       & 59964.867                       & 78.787$\%$                                                                                &  299.359                                                      \\
\multicolumn{1}{c}{PPO}   & -   & -   & -  &  56     & 30.961               &   71204.500                                                      &  \textbf{99.451}\bm{$\%$}                                                                               & 82.139                                                       \\ \bottomrule
\end{tabular}

\begin{tablenotes}    
                      
\item[1] A: Conformer, B: DDPG-based Power Flow Optimization Algorithm, C: Dispatching Necessity Evaluation Mechanism
\end{tablenotes}            
    \end{threeparttable} 
\end{center}
\end{table*}

\begin{figure*}[t]
    \centering
    \centerline{\includegraphics[width=\textwidth,height=4.7cm]{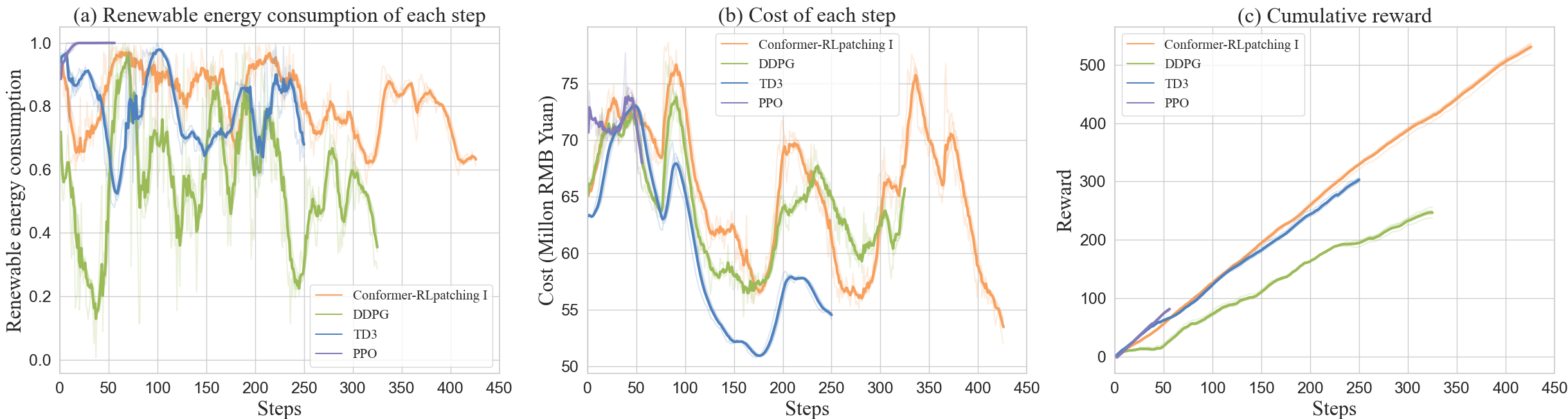}}
    \caption{Simulation Results of Each Step}
    \label{step_result}
\end{figure*}

\subsection{Performance of Conformer-RLpatching With Different Hyper Parameters and Ablation Experiments}
According to Section \ref{prediction result}, the length of historical data input to Conformer-RLpatching is set as 56 with the lowest prediction error. Conformer outputs the prediction for the next 10 days in all experiments except for Conformer-RLpatching II, which only provides prediction for the next day. 

In order to compare the impact of long-term and short-term renewable energy prediction on dispatching effect, this paper designs a comparative experiment between Conformer-RLpatching I and II. As shown in TABLE \ref{SiR}, Conformer-RLpatching I enables the simulator to run 117 more steps than Conformer-RLpatching II whilst satisfying the constrains, and the total reward of Conformer-RLpatching I is 52.352$\%$ higher.

Conformer-RLpatching I and III-VIII test the impact of different parameters in the dispatching necessity evaluation mechanism, including ${\mathop \varphi \nolimits_R}$, ${\mathop \varphi \nolimits_L}$, $\mathop n\nolimits_{dis}$, on active power flow dispatching of hybrid power grid. According to TABLE \ref{SiR}, overall, when ${\mathop \varphi \nolimits_R}$ and ${\mathop \varphi \nolimits_L}$ are both 0.8 and $\mathop n\nolimits_{dis}$ is 40, Conformer-RLpatching performs best, and its total reward reaches 527. When ${\mathop \varphi \nolimits_R}$ changes, the security score and renewable energy utilization rate show a downward trend, but the average cost of each step changes little. For example, when ${\mathop \varphi \nolimits_R}$ increases from 0.8 to 0.9, the utilization rate of renewable energy decreases by 4.499$\%$. When the critical value of 
branch current ${\mathop \varphi \nolimits_L}$ increases, the security of power grid is seriously affected. As ${\mathop \varphi \nolimits_L}$ increases from 0.8 to 0.9, the security score decreases by 7.429$\%$. As $\mathop n\nolimits_{dis}$ rises to 40, the stability of the power grid reaches a peak. When $\mathop n\nolimits_{dis}$ increases from 40 to 45, the number of operation steps and security score decrease by 2.113$\%$ and 5.843$\%$, respectively, despite the utilization rate of renewable energy increasing by 1.348$\%$.

In addition, ablation experiments are carried out to explore the separate contributions of Conformer and dispatching necessity evaluation mechanism. It can be seen that the security score decreases by 3.952$\%$ without C. This indicates that the dispatching necessity evaluation mechanism can ensure the stable operation of the power grid system. Besides, the total reward significantly reduces by 54.190$\%$ discarding A, which illustrates that Conformer can assist the grid dispatching and abate the negative impact of renewable energy fluctuations.

\subsection{Comparison With Other Dispatching Methods}
This paper compares Conformer-RLpatching with DDPG, distributed classification replay twin delayed deep deterministic policy gradient (DCR-TD3) \cite{e3} and proximal policy optimization (PPO) to further verify the dispatching effect of Conformer-RLpatching. The simulation results are shown in TABLE \ref{SiR}. The security score of Conformer-RLpatching is 25.758$\%$ and 63.319$\%$ higher than that of DDPG and DCR-TD3. The total reward is improved by 76.149$\%$ compared with DCR-TD3 having the second best performance. The renewable energy utilization rate of Conformer-RLpatching is 4.540$\%$ higher than that of other methods on average.

More intuitively, Fig. \ref{step_result} depicts the average renewable energy utilization rate, the average cost, and the cumulative reward of each step of the above dispatching methods in an epoch. As shown in Fig. \ref{step_result} (a), despite PPO reaching relatively high renewable energy utilization rate in the beginning, its dispatching is highly unstable with an end at only step 50. In contrast, Conformer-RLpatching, which enables superior stability and efficient renewable energy utilization rate, is noticeably the best performer on the renewable energy utilization issue, albeit at the cost of slightly high cost as presented in Fig. \ref{step_result} (b). Fig. \ref{step_result} (c) shows the overall performance in dispatching. The rewards obtained by PPO and DCR-TD3 improve fast initially whereas coming to an end quickly, which indicates their instability. DDPG runs relatively more steps than PPO and DCR-TD3, but it obtains the lowest rewards of them in the epoch. To sum up, Conformer-RLpatching achieves the most running steps and the highest cumulative reward, verifying its superiority among the state-of-the-art dispatching methods.

\begin{table}[H]
\caption{Comparison With the Top Three Teams}
\label{comptition}
\centering
\scalebox{1.1}[1.1]{
\begin{tabular}{cc}
\toprule
                     & Total Reward \\ \hline
Conformer-RLpatching I & \textbf{527.318}      \\
The first ranked team       & 510.091      \\
The second ranked team      & 509.206      \\
The third ranked team       & 507.239      \\
\bottomrule
\end{tabular}
}
\end{table}
TABLE \ref{comptition} compares the total reward of Conformer-RLpatching 
and the top three teams in State Grid Dispatching AI Innovation Competition. The results show that Conformer-RLpatching achieves a remarkable total reward up to 527.318 superior to 510.091 obtained by the best team.

\section{Conclusion}\label{conclu}

The emerging AI technology is gradually integrated into the security constrained economic dispatching of the hybrid energy grid. This paper proposes a Conformer-RLpatching to achieve multi-objective dispatching under the long-term fluctuations of renewable energy. Equipped with the confidence estimation method, the Conformer can provide accurate long-term renewable energy prediction to reduce the impact of new energy uncertainty on dispatching. What’s more, this paper designs RLpatching to realize active power flow optimization and puts forward a dispatching necessity evaluation mechanism to reduce unnecessary dispatching and improve the stability of the power grid. Extensive experiments based on the SG-126 power grid simulator demonstrate that the proposed Conformer-RLpatching is superior to other state-of-the-art methods. The future work will consider the impact of additional factors, such as weather and social activities, on renewable energy to improve the prediction accuracy, and further study the multi-objective dispatching of hybrid power grid under source-load side uncertainty.

\appendices
\section{Prediction Results}
\label{App}
As for prediction models, historical data of the past 48 days, 56
days, 64 days, 72 days, 80 days, 88 days, and 96 days are fed
into models separately, and the length of prediction
is set as 15, 20, 25 and 30. All experimental results are shown in the following table. 
\begin{table}[H]
\resizebox{\columnwidth}{30mm}{
\begin{tabular}{cccccccc}
\toprule
\begin{tabular}[c]{@{}c@{}}Input data\\ length\end{tabular} & \begin{tabular}[c]{@{}c@{}}Output data\\ length\end{tabular} & Conformer & Informer & LSTMa   & Prophet & CNN    & \begin{tabular}[c]{@{}c@{}}CNN-\\ LSTM\end{tabular} \\ \hline
\multirow{4}{*}{48}                                         & 15                                                           & 5.0648    & 5.937    & 10.9698 & 8.9671  & 6.6045 & 7.8669                                              \\
                                                            & 20                                                           & 5.3919    & 6.7637   & 11.3034 & 9.7796  & 7.5408 & 8.6973                                              \\
                                                            & 25                                                           & 5.8599    & 7.2799   & 12.1275 & 10.8414 & 8.0617 & 9.2094                                              \\
                                                            & 30                                                           & 6.2192    & 7.8346   & 12.225  & 11.8916 & 8.5013 & 9.5646                                              \\ \hline
\multirow{4}{*}{56}                                         & 15                                                           & 4.9995    & 5.9677   & 10.864  & 9.4211  & 6.7521 & 8.2419                                              \\
                                                            & 20                                                           & 5.2718    & 6.683    & 12.7792 & 10.3752 & 7.4893 & 9.3553                                              \\
                                                            & 25                                                           & 5.9732    & 7.3198   & 13.0704 & 11.2745 & 8.0038 & 9.692                                               \\
                                                            & 30                                                           & 6.178     & 7.7981   & 12.5671 & 12.3669 & 8.572  & 9.7102                                              \\ \hline
\multirow{4}{*}{64}                                         & 15                                                           & 5.0076    & 6.0773   & 11.1334 & 9.5111  & 6.8096 & 8.3122                                              \\
                                                            & 20                                                           & 5.3985    & 6.7428   & 12.8136 & 10.3964 & 7.5571 & 9.4028                                              \\
                                                            & 25                                                           & 6.0351    & 7.3137   & 13.0573 & 11.2522 & 7.9928 & 9.4484                                              \\
                                                            & 30                                                           & 6.1866    & 7.7993   & 12.923  & 12.5742 & 8.6852 & 9.9086                                              \\ \hline
\multirow{4}{*}{72}                                         & 15                                                           & 5.0214    & 6.081    & 11.1267 & 9.5745  & 6.8373 & 8.3324                                              \\
                                                            & 20                                                           & 5.4412    & 6.7233   & 12.9052 & 10.4459 & 7.5659 & 9.3964                                              \\
                                                            & 25                                                           & 5.906     & 7.3482   & 13.1027 & 11.3085 & 8.096  & 9.5547                                              \\
                                                            & 30                                                           & 6.2055    & 7.8026   & 12.9616 & 12.5302 & 8.6734 & 9.9164                                              \\ \hline
\multirow{4}{*}{80}                                         & 15                                                           & 5.1835    & 6.0975   & 11.3142 & 9.6998  & 6.8945 & 8.4128                                              \\
                                                            & 20                                                           & 5.5034    & 6.7255   & 12.8648 & 10.4943 & 7.5203 & 9.4645                                              \\
                                                            & 25                                                           & 5.9607    & 7.3541   & 12.9361 & 11.3496 & 8.1734 & 9.6077                                              \\
                                                            & 30                                                           & 6.2565    & 7.844    & 13.1125 & 12.5463 & 8.7174 & 9.8437                                              \\
\bottomrule
\end{tabular}}
\end{table}

\begin{table}[H]
\resizebox{\columnwidth}{15mm}{
\begin{tabular}{cccccccc}
\toprule
\begin{tabular}[c]{@{}c@{}}Input data\\ length\end{tabular} & \begin{tabular}[c]{@{}c@{}}Output data\\ length\end{tabular} & Conformer & Informer & LSTMa   & Prophet & CNN    & \begin{tabular}[c]{@{}c@{}}CNN-\\ LSTM\end{tabular} \\ \hline

\multirow{4}{*}{88}                                         & 15                                                           & 4.9972    & 6.1014   & 11.3667 & 9.6289  & 6.8981 & 8.3959                                              \\
                                                            & 20                                                           & 5.4511    & 6.6902   & 12.8955 & 10.5025 & 7.6697 & 9.773                                               \\
                                                            & 25                                                           & 5.9611    & 7.3878   & 13.0564 & 11.5041 & 8.3747 & 9.7416                                              \\
                                                            & 30                                                           & 6.2532    & 7.8635   & 13.2063 & 12.5195 & 8.7748 & 9.9718                                              \\ \hline
\multirow{4}{*}{96}                                         & 15                                                           & 5.1985    & 6.1397   & 11.4291 & 9.8016  & 6.9034 & 8.4526                                              \\
                                                            & 20                                                           & 5.5323    & 6.7316   & 13.0103 & 10.6395 & 7.6748 & 9.8253                                              \\
                                                            & 25                                                           & 5.9236    & 7.3796   & 13.1256 & 11.8777 & 8.4523 & 9.8562                                              \\
                                                            & 30                                                           & 6.3164    & 7.8826   & 13.1813 & 12.6051 & 8.8516 & 9.9868                                              \\ \bottomrule
\end{tabular}
}
\end{table}


\bibliographystyle{IEEEtran}
\bibliography{cite_MPCE}

\text{}\\
\noindent\textbf{Xinhang Li} received the B.E. degree in communication engineering from Beijing University of Posts and Telecommunications (BUPT), Beijing, China, in 2021. He is currently pursuing the Ph.D. degree in information and communication engineering from the School of Artificial Intelligence, BUPT. His research interests include deep reinforcement learning, optimal power flow and intelligent information processing.\\

\noindent\textbf{Zihao Li} is currently pursuing the B.E. degree in telecommunications engineering with management with Beijing University of Posts and Telecommunications, Beijing, China. His current research interest includes machine learning and artificial intelligence.\\

\noindent\textbf{Nan Yang} received the B.S. and M.S. degrees in electrical engineering from Beijing Institute of Technology (BIT), Beijing, China, in 2015 and 2018, respectively. She works for China Electric Power Research Institute, and her research interests include big data analysis and artificial intelligence application in the field of power dispatching automation.\\

\noindent\textbf{Zheng Yuan} received the B.E. degree in information engineering from Beijing University of Posts and Telecommunications (BUPT), Beijing, China, in 2021. He is currently pursuing an M.S. degree from the School of Artificial Intelligence, BUPT. His research interests are reinforcement learning, power dispatching and intelligent information processing.\\

\noindent\textbf{Qinwen Wang} received the B.E. degree in digital media technology from Communication University of China (CUC), Beijing, China, in 2020. She is currently pursuing an M.S. degree from the School of Artificial Intelligence, Beijing University of Posts and Telecommunications (BUPT), Beijing, China. Her research interests include reinforcement learning, smart grid and cooperative intelligent transportation systems.\\

\noindent\textbf{Yiying Yang} received the B.E. degree in communication engineering from Beijing University of Posts and Telecommunications (BUPT), Beijing, China, in 2021. She is currently pursuing the M.S. degree in information and communication engineering with BUPT. Her research interests include reinforcement learning, power dispatching and cooperative connected vehicles control.\\

\noindent\textbf{Yupeng Huang} received the B.S. degree in electrical engineering from Tsinghua University (THU), Beijing, China, in 2016, and received the M.S. degree in electric power system and automation from China Electric Power Research Institute (CEPRI), Beijing, China, in 2019.  He works for China Electric Power Research Institute and his research interests include power dispatching and automation.\\

\noindent\textbf{Xuri Song} received the B.E. and M.S. degrees in electrical engineering from China Agricultural University, Beijing, China, in 2009 and 2011. He works for China Electric Power Research Institute. His research interests include power grid analysis and artificial intelligence application.\\

\noindent\textbf{Lei Li} is currently an Associate Professor with the School of Artificial Intelligence, Beijing University of Posts and Telecommunications, China. Her research interests include intelligent information processing, deep learning, machine learning, and natural language processing.\\

\noindent\textbf{Lin Zhang} (Member, IEEE) received the B.S. and Ph.D. degrees from the Beijing University of Posts and Telecommunications (BUPT), Beijing, China, in 1996 and 2001, respectively. He is currently the Director of Beijing Bigdata Center and also a Professor of BUPT. He was a Postdoctoral Researcher with Information and Communications University, South Korea. He used to hold a Research Fellow position with Nanyang Technological University, Singapore. In 2004, he joined BUPT as a Lecturer, then an Associate Professor in 2005, and a Professor in 2011. He has authored more than 120 papers in referenced journals and international conferences. His research interests include intelligent information processing, deep learning, mobile cloud computing and Internet of Things.

\end{document}